\documentclass[preprint]{article}
\usepackage{graphicx} 
\usepackage[nonanonymous]{neurips_2026}

\usepackage{algorithm} 
\usepackage{algpseudocode}
\usepackage{amsmath}
\usepackage{tikz}
\usepackage{url}
\usepackage{booktabs}
\usepackage[most]{tcolorbox}
\usepackage{pgfplots}
\usepackage{listings}
\usepackage{xcolor}
 \usepackage{amssymb}
 
\definecolor{promptbg}{gray}{0.975}
\definecolor{promptframe}{gray}{0.68}
\definecolor{prompttitle}{gray}{0.12}

\lstdefinestyle{promptstyle}{
  basicstyle=\ttfamily\scriptsize,
  breaklines=true,
  breakatwhitespace=false,
  columns=fullflexible,
  keepspaces=true,
  showstringspaces=false,
  tabsize=2,
  aboveskip=0pt,
  belowskip=0pt
}

\newtcblisting{promptbox}[1]{
  enhanced,
  breakable,
  listing only,
  listing engine=listings,
  listing options={style=promptstyle},
  title={#1},
  colback=promptbg,
  colframe=promptframe,
  coltitle=prompttitle,
  fonttitle=\bfseries\small,
  boxrule=0.55pt,
  arc=1.2mm,
  outer arc=1.2mm,
  left=1.8mm,
  right=1.8mm,
  top=1.2mm,
  bottom=1.2mm,
  before skip=8pt,
  after skip=10pt
}

\usetikzlibrary{calc,backgrounds,arrows.meta,fit,positioning,shapes.geometric}

\pgfdeclarelayer{bg}
\pgfsetlayers{bg,main}

\definecolor{semcol}{RGB}{124,32,114}     
\definecolor{epicol}{RGB}{0,150,170}      
\definecolor{epifill}{RGB}{198,238,245}   
\definecolor{bgpurple}{RGB}{244,236,250}  
\definecolor{qorange}{RGB}{225,135,35}    
\definecolor{qfill}{RGB}{255,238,212}     
\definecolor{tracecol}{RGB}{20,30,200}    
\definecolor{glow}{RGB}{255,205,110}

\title{Selective Forgetting: A Graph-Based Memory Framework for Long-Term LLM Agents}

\author{
Theo Rusu \\
Department of Computer Science \\
Toronto Metropolitan University \\
Toronto, Ontario, Canada \\
\texttt{trusu@torontomu.ca}
\And
Sourena Khanzadeh\thanks{This work was conducted prior to the author's affiliations with Flybits and The Creative School. The author was previously affiliated with the Department of Computer Science at Toronto Metropolitan University.} \\
The Creative School \\
Toronto Metropolitan University \\
Toronto, Ontario, Canada \\
Flybits, Creative AI Hub \\
Toronto, Ontario, Canada \\
\texttt{sourena.khanzadeh@torontomu.ca}
\And
Manar Alalfi \\
Department of Computer Science \\
Toronto Metropolitan University \\
Toronto, Ontario, Canada \\
\texttt{manar.alalfi@torontomu.ca}
}

\begin{document}

\maketitle

\begin{abstract}
Knowledge graphs have been proposed as a structured alternative to flat
  retrieval-augmented generation for long-term agent memory, on the assumption
  that representing conversations as entities and relations improves recall. We
  evaluate that assumption directly. Our framework extracts each conversational
  turn into typed nodes and attributed edges, answers questions from a two-hop
  subgraph, and periodically prunes nodes that score low on a weighted
  combination of recency, access frequency, degree centrality, and age. On
  LongMemEval, the graph does not outperform a flat vector baseline at a matched
  candidate-generation budget of five retrieval roots: token F1 is $0.417$
  against $0.468$, and a paired bootstrap over 500 questions gives
  $\Delta = -0.050$ (95\% CI $[-0.085, -0.016]$). The gap is widest on questions
  that require recalling a specific prior assistant turn, where judged
  correctness falls from $0.911$ to $0.607$, suggesting that decomposing a turn
  into entities discards the surface form these questions depend on. The
  forgetting module is more successful. Applied once to a persistent
  27{,}021-node graph, it removes 9.8\% of nodes and 9.5\% of stored bytes;
  token F1 is unchanged ($+0.001$, 95\% CI $[-0.015, +0.016]$) and judged
  correctness falls by $1.6$ points, with the 95\% interval bounding any loss at
  $3.8$ points ($[-0.038, +0.006]$). Because our extractor is a single small
  model evaluated on one benchmark, these results characterise this
  extraction-based pipeline rather than graph-structured memory in general.

Code: \url{https://github.com/skhanzad/Selective-Amnesia}
\end{abstract}

\section{Introduction}

Large Language Models have rapidly evolved from standalone text generators into the foundation of complex, agentic systems that are able to reason, plan, and use external tools. These systems are increasingly being adopted as personal AI assistants that interact with users over prolonged periods of time. The quality of these interactions is reliant not only on the model's immediate reasoning ability, but also on the model's capacity to incorporate information from past exchanges. As such, memory becomes a critical component of these systems and is typically classified into two subcategories: short-term memory and long-term memory.

The current widely adopted long-term memory approach is retrieval-augmented generation \cite{lewis2020retrieval}, where dense vector stores are indexed and queried at inference time to retrieve relevant entries that augment the model's output. This method models memory as a flat, similarity-based retrieval system, which has been shown to be sensitive to noise, prone to retrieving irrelevant or redundant context, and is limited in its ability to support multi-hop reasoning or maintain coherent long-term knowledge \cite{gao2023retrieval}.

To address these limitations, recent work has explored structured memory representations based on knowledge graphs, where information is organized as entities and their relationships rather than independent embeddings. In these systems, memories are encoded as nodes and edges, allowing more complex semantic and relational representations to be stored \cite{ji2022survey, peng2023knowledge}.

However, existing graph-based approaches mainly focus on how information is added and maintained for consistency \cite{chhikara2025mem0}. This does not address a fundamental challenge of long-term memory systems: unbounded growth. As interactions accumulate over time, memory stores become increasingly large, creating a range of downstream negative effects, including degraded retrieval quality, higher computational cost, and the retention of low-utility information.

As memory accumulates over prolonged interactions, the system must integrate new information and manage the relevance of existing knowledge. Prior work in continual learning and neural memory systems has shown that effective memory requires mechanisms for selective retention and forgetting, as retaining all information leads to performance degradation \cite{kirkpatrick2017overcoming, wei2026fademem}.

In this work, we investigate whether structuring long-term conversational memory as a knowledge graph meaningfully improves retrieval and reasoning in LLM agents, and whether systems can remain efficient over extended interactions. We introduce a graph-based memory framework with an explicit forgetting module that controls the lifecycle of stored information. Rather than assuming structural representations are uniformly beneficial, our study empirically characterizes where graph-based memory helps and where it degrades performance. In addition, we show that selective forgetting based on recency, frequency, and structural importance can reduce memory size without materially affecting retrieval quality. These findings highlight that effective long-term memory requires structured representation and careful design of update and retention mechanisms.


\section{Related Work}

\subsection{Memory in LLM-based Agents}

Equipping neural systems with explicit memory predates the current generation of
language models. Early differentiable architectures such as Neural Turing
Machines~\cite{graves2014neural} and End-to-End Memory Networks~\cite{sukhbaatar2015end}
coupled a controller with an addressable external store, establishing the
read/write abstraction that later memory systems inherit. As LLMs became the
backbone of agentic systems, memory was repurposed to persist information between
turns and sessions rather than within a single forward pass~\cite{zhang2024survey}.
A common design is the memory stream of Generative Agents~\cite{park2023generative},
which logs observations and retrieves them using a combination of recency, importance,
and relevance, and periodically synthesizes higher-level reflections. Subsequent
systems extend this idea along different axes: MemoryBank~\cite{zhong2024memorybank}
introduces an updating scheme inspired by the Ebbinghaus forgetting
curve~\cite{ebbinghaus1885memory}; MemGPT~\cite{packer2024memgpt} treats memory as an
operating-system-style hierarchy that pages information between a bounded context
window and external storage; ReadAgent~\cite{lee2024readagent} compresses very long
contexts into gist memories; and Think-in-Memory~\cite{liu2023think},
Self-Controlled Memory~\cite{wang2024selfcontrolled}, and MemLLM~\cite{modarressi2024memllm}
give the model explicit control over what is stored and recalled. These approaches
also enable long-term dialogue settings studied by~\cite{xu2022beyond}.
Most of this line of work, however, organizes memory as a flat collection of
entries and emphasizes writing and reading rather than principled removal.

\subsection{Retrieval-Augmented Generation}

The dominant strategy for grounding LLM outputs in external knowledge is
retrieval-augmented generation~\cite{lewis2020retrieval}, which retrieves relevant
passages from a non-parametric store and conditions generation on them. Dense
retrieval~\cite{karpukhin2020dense} and jointly pre-trained retrieval-reading
models such as REALM~\cite{guu2020realm} and RETRO~\cite{borgeaud2022improving}
improved retrieval quality on scale, while retrieval has been shown to reduce
hallucinations in dialogue~\cite{shuster2021retrieval}. More recent variants add
self-reflective control over when and what to retrieve~\cite{asai2024selfrag} and
specialize models for conversational settings~\cite{luo2024chatqa}. However, as surveyed
by~\cite{gao2023retrieval}, RAG fundamentally models memory as a flat,
similarity-based lookup over independent embeddings. This makes it sensitive to
retrieval noise and redundancy and limits its capacity for multi-hop reasoning or
maintaining coherent long-term knowledge, motivating more structured
representations of memory.

\subsection{Graph-Structured Memory}

Knowledge graphs offer a structured alternative in which information is represented
as entities and the relations between them~\cite{ji2022survey,peng2023knowledge}.
A growing body of work integrates such structure with LLMs~\cite{pan2024unifying},
ranging from prompting with retrieved triples~\cite{baek2023knowledge} to letting
the model reason by traversing a graph~\cite{sun2024thinkongraph}. For
retrieval specifically, GraphRAG~\cite{edge2024local} constructs an entity graph
and community summaries to support query-focused summarization, and
HippoRAG~\cite{gutierrez2024hipporag} draws on hippocampal indexing theory to
combine a knowledge graph with graph-based retrieval for long-term recall. In the
agent-memory setting, systems such as Mem0~\cite{chhikara2025mem0} adopt graph
representations to store and consolidate user information across sessions. These
methods demonstrate the benefits of relational structure for retrieval and
reasoning, but they concentrate on how information is added and kept consistent and
largely leave unbounded growth of the memory store unaddressed.

\subsection{Forgetting and Memory Retention}

The need to forget is well established outside of agent memory. In human cognition,
retention decays predictably over time~\cite{ebbinghaus1885memory}. In neural
networks, naive sequential learning induces catastrophic
forgetting~\cite{mccloskey1989catastrophic}, prompting mechanisms that protect
important parameters~\cite{kirkpatrick2017overcoming}; the broader phenomenon of
forgetting in deep learning is surveyed by~\cite{wang2024forgetting}, and machine
unlearning studies the deliberate removal of specific
information~\cite{bourtoule2021machine}. A consistent finding across these areas is
that effective memory requires selective retention rather than indefinite
accumulation. This principle has only recently been applied to agent memory:
FadeMem~\cite{wei2026fademem} introduces biologically inspired forgetting to keep
agent memory efficient. Our work is closest in spirit to this direction, but
couples forgetting with a graph-structured store: rather than treating retention as
a post-hoc filter over flat entries, we integrate a forgetting module into the
life cycle of nodes and edges, so that obsolete or low-utility memories are removed
while relational structure is preserved.

\subsection{Evaluating Long-Term Memory}

Assessing memory over extended interactions requires dedicated benchmarks. 
LoCoMo~\cite{maharana2024evaluating} evaluates very long-term conversational memory,
and LongMemEval~\cite{wu2025longmemeval} probes chat assistants on long-term
interactive memory abilities such as multi-session reasoning and knowledge updates.
We adopt LongMemEval to evaluate whether structured memory with forgetting
sustains high-quality recall as interactions accumulate.

\section{Methodology}
This study proposes a graph-based conversational memory framework that models
interactions as a structured, evolving knowledge graph. Rather than storing
past exchanges as independent embeddings, the system maintains entities and
their relationships as nodes and edges, continuously updates this
graph as new conversational turns arrive, and periodically prunes low-importance
nodes to bound graph growth over long interactions.

\subsection{Architecture}

\begin{figure*}[!t]
\centering
\resizebox{\textwidth}{!}{%
\begin{tikzpicture}[
    font=\small\sffamily,
    node distance=3.5mm and 30mm,
    blk/.style={
        rectangle, rounded corners=2pt, draw=black!70, semithick,
        fill=white, minimum width=33mm, minimum height=7mm,
        align=center, inner sep=2pt
    },
    blkA/.style={blk, fill=blue!8},
    blkB/.style={blk, fill=green!9},
    blkC/.style={blk, fill=red!8},
    io/.style={
        rectangle, rounded corners=2pt, draw=black!70, semithick, dashed,
        fill=black!3, minimum width=33mm, minimum height=7mm,
        align=center, inner sep=2pt
    },
    kg/.style={
        cylinder, shape border rotate=90, draw=black, thick,
        aspect=0.16, fill=yellow!15, minimum width=44mm,
        minimum height=17mm, align=center
    },
    flow/.style={-{Stealth[length=2.2mm]}, semithick, black!75},
    kgflow/.style={-{Stealth[length=2.8mm]}, thick},
    stagebox/.style={
        rectangle, rounded corners=4pt, draw=black!40, semithick,
        inner sep=4mm
    },
    stagetitle/.style={font=\small\sffamily\bfseries, anchor=south west}
]
 
\node[io]                          (query)   {User question};
\node[blkA, below=of query]        (extract1){Entity extraction};
\node[blkA, below=of extract1]     (embed1)  {Descriptor embedding};
\node[blkA, below=of embed1]       (search)  {Top-5 node search\\{\scriptsize cosine $> 0.75$}};
\node[blkA, below=of search]       (traverse){2-hop subgraph traversal\\{\scriptsize $\leq 15$ nodes}};
\node[io,   below=of traverse]     (answer)  {LLM answer};
 
\draw[flow] (query)    -- (extract1);
\draw[flow] (extract1) -- (embed1);
\draw[flow] (embed1)   -- (search);
\draw[flow] (search)   -- (traverse);
\draw[flow] (traverse) -- (answer);
 
\begin{pgfonlayer}{bg}
\node[stagebox, fit=(query)(answer), fill=blue!3] (stage1) {};
\end{pgfonlayer}
\node[stagetitle] at (stage1.north west) {1.\ Retrieval};
 
\node[io, right=of query]          (turn)    {Conversational turn};
\node[blkB, below=of turn]         (extract2){LLM node \& edge\\extraction};
\node[blkB, below=of extract2]     (embed2)  {Descriptor embedding};
\node[blkB, below=of embed2]       (dedup)   {De-duplication\\{\scriptsize title index, cosine $> 0.92$}};
\node[blkB, below=of dedup]        (write)   {Write new /\\merge existing};
 
\draw[flow] (turn)     -- (extract2);
\draw[flow] (extract2) -- (embed2);
\draw[flow] (embed2)   -- (dedup);
\draw[flow] (dedup)    -- (write);
 
\begin{pgfonlayer}{bg}
\node[stagebox, fit=(turn)(write), fill=green!4] (stage2) {};
\end{pgfonlayer}
\node[stagetitle] at (stage2.north west) {2.\ Update};
 
\node[io, right=of turn]           (trigger) {Every 400 turns};
\node[blkC, below=of trigger]      (score)   {Importance scoring\\
    {\scriptsize recency $\cdot$ frequency $\cdot$}\\
    {\scriptsize centrality $\cdot$ turn age}};
\node[blkC, below=of score]        (thresh)  {Threshold check};
\node[blkC, below=of thresh]       (prune)   {Prune nodes \&\\incident edges};
 
\draw[flow] (trigger) -- (score);
\draw[flow] (score)   -- (thresh);
\draw[flow] (thresh)  -- node[right, font=\scriptsize] {$s < 0.10$} (prune);
 
\begin{pgfonlayer}{bg}
\node[stagebox, fit=(trigger)(prune), fill=red!3] (stage3) {};
\end{pgfonlayer}
\node[stagetitle] at (stage3.north west) {3.\ Retention};
 
\node[kg] (kgnode) at ($(stage2.south)+(0,-21mm)$)
    {Persistent knowledge graph\\
     {\scriptsize typed nodes $\cdot$ attributed edges $\cdot$ vector index}};
 
\draw[kgflow, draw=blue!60!black]
    (kgnode.west) .. controls +(-18mm,0) and +(0,-12mm) .. (stage1.south)
    node[pos=0.4, below left, font=\scriptsize\itshape, text=blue!60!black]
    {read: nodes \& subgraphs};
 
\draw[kgflow, draw=green!45!black]
    (stage2.south) -- (kgnode.north)
    node[midway, right, font=\scriptsize\itshape, text=green!45!black]
    {write / merge};
 
\draw[kgflow, draw=red!60!black]
    (stage3.south) .. controls +(0,-12mm) and +(18mm,0) .. (kgnode.east)
    node[pos=0.5, below right, font=\scriptsize\itshape, text=red!60!black]
    {prune};
 
\draw[kgflow, dashed, draw=black!50]
    (kgnode.east) .. controls +(20mm,3mm) and +(11mm,-23mm) .. (score.east)
    node[pos=0.45, right, font=\scriptsize\itshape, text=black!60]
    {graph statistics};
 
\end{tikzpicture}
}
\caption{Overview of the proposed memory framework. Each conversational
turn is processed by the \emph{update} stage, which extracts, embeds,
de-duplicates, and writes nodes and edges into the persistent knowledge
graph. At question time, the \emph{retrieval} stage embeds the referenced
entities, selects the top-5 matching nodes ranked by cosine similarity,
and expands them via a 2-hop subgraph traversal to build the
answer-generation context. Every 400 turns, the \emph{retention} stage
scores each node by recency, access frequency, centrality, and turn age,
pruning nodes whose importance falls below the threshold together with
their incident edges.}
\label{fig:pipeline}
\end{figure*}
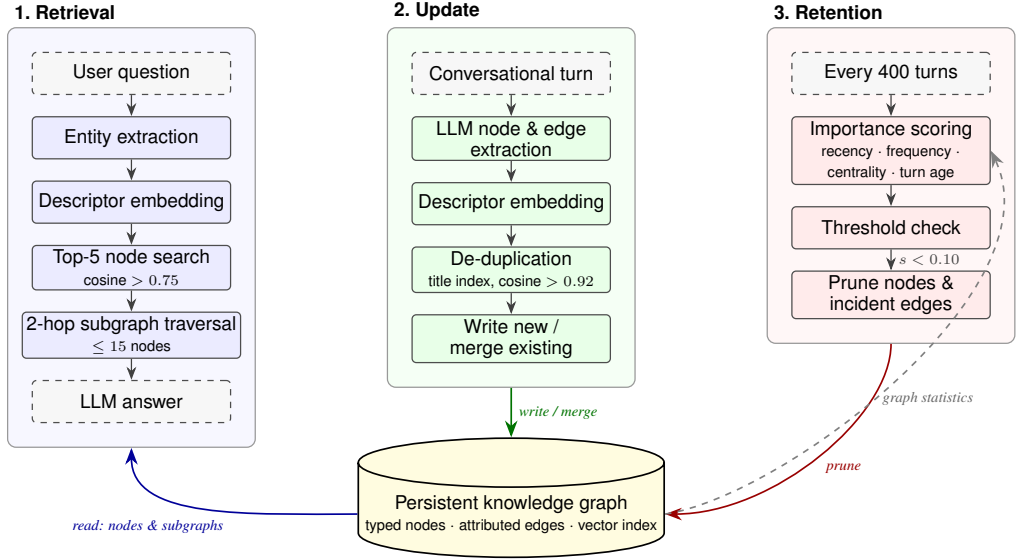

The framework is organized as a three-stage pipeline: \emph{retrieval}, \emph{update},
and \emph{retention}. In the \emph{retrieval} stage, the subgraph most relevant to the
current question is selected and serialized as context for answer generation. In the
\emph{update} stage, an LLM extracts entities and relationships from the current
conversational turn and integrates them into the knowledge graph. In the \emph{retention}
stage, a forgetting module scores every stored node and removes those whose importance
falls below a threshold, bounding graph growth and discarding low-utility information.
Figure~\ref{fig:pipeline} gives an overview of the full pipeline and how the three
stages interact with the persistent knowledge graph.

\subsection{Knowledge Graph Schema}

The graph consists of typed nodes and edges with both types and attributes. Each node
carries a \texttt{label}, a short \texttt{title}, a natural-language
\texttt{content} description, a flat \texttt{attributes} dictionary, and
system fields for temporal, access, and retention tracking (\texttt{created\_at},
\texttt{access\_count}, \texttt{last\_accessed\_at},
\texttt{turns\_at\_creation}, \texttt{importance\_score}).

Nine node labels are defined in a fixed ontology: \textit{Person},
\textit{Organization}, \textit{Location}, \textit{Event}, \textit{Concept},
\textit{Artifact}, \textit{Preference}, \textit{Goal}, and \textit{Skill}.
Edges connect pairs of nodes via a typed \texttt{relationship} predicate and
may carry their own \texttt{attributes} dictionary to capture relational
properties such as duration, confidence, or quantity without introducing
additional nodes.

\subsection{Extraction Pipeline}

Each conversational turn is processed by a single LLM extraction call (GPT-4o-mini)
using a structured system prompt that defines the ontology, output schema,
and extraction rules. Turns are prefixed with their speaker role (\texttt{[Role: user]}
or \texttt{[Role: assistant]}) so the extractor handles each appropriately.

User and assistant turns are processed differently during extraction. For user turns, the system captures facts about the user, including their preferences, goals, skills, and relationships. Assistant turns are processed in two modes: (a) user-related facts that the assistant references or confirms, and (b) factual claims, recommendations, and named-entity information stated by the assistant, enabling the system to recall information provided in prior turns.


Turns that contain only superfluous or generic
filler words produce empty output.

The extractor returns strict JSON conforming to the graph schema. A validation
layer rejects any output that fails to parse, references undefined ontology
types, or contains structurally invalid node-edge references.

\subsection{Embedding and De-duplication}

Node descriptors are constructed from each node's label, title, and content and
embedded using \texttt{nomic-embed-text} served locally via Ollama. The resulting
vectors are stored alongside each node in the graph's \texttt{vector\_index}
dictionary and persisted with the graph JSON file.

Before a new node is written to the graph, two de-duplication checks are applied in sequence.
The first is title-based: if an existing node shares the same label and normalized title
as the incoming node, it is identified as a match and the embedding check is skipped entirely.
To enable the title-based check, the system maintains a \texttt{title\_index} - a dictionary
mapping each node's normalized title to its node ID — which is updated whenever a new node is
created, allowing O(1) lookup without any vector computation. The second is embedding-based,
applied only when the title check finds no match: the incoming node's vector is compared to all
stored vectors via linear cosine scan; if the nearest existing node has a cosine similarity above
0.92, it is treated as the same entity. When either check identifies a match, the matched node's
\texttt{access\_count} and \texttt{last\_accessed\_at} fields are updated, providing a frequency signal
to the importance scoring module before formal retrieval occurs.

\subsection{Subgraph Retrieval}

At inference time, a lightweight extraction call identifies the entities referenced by the current
question and embeds their descriptors. 
An exhaustive cosine-similarity search is performed over all stored node embeddings. Nodes with a similarity score above 0.75 are retained and ranked by score. We use breadth-first search (BFS) starting from the top-5 retrieved nodes, expanding the graph up to two hops from these root nodes and stopping once a maximum of 15 nodes has been collected. Justifications for these parameter choices are provided in Appendix~\ref{appendix:just}.

The retrieved nodes and edges are serialized as a compact textual representation and included into the answer-generation prompt as contextual memory.

Visited nodes in a retrieved subgraph has its \texttt{access\_count} incremented and 
\texttt{last\_accessed\_at} updated to the current session timestamp.

\subsection{Importance Scoring and Forgetting}

Each node is assigned an importance score that combines four components:

\begin{equation*}
\begin{aligned}
  \text{Score} ={}\ & w_r \cdot \text{recency}(t) + w_f \cdot \text{frequency}(c) \\
                    & + w_c \cdot \text{centrality}(d) + w_t \cdot \text{turns\_decay}(k),
\end{aligned}
\end{equation*}

\noindent where \emph{recency} is an exponential decay from the node's last
access timestamp with a 90-day half-life; \emph{frequency} is the
log-normalized access count; \emph{centrality} is the log-scaled edge degree;
and \emph{turns\_decay} is an exponential decay from the node's creation turn
with a 1{,}000-turn half-life. The weights are set to $w_r = 0.35$,
$w_f = 0.25$, $w_c = 0.20$, $w_t = 0.20$.

The forgetting module is invoked every 400 conversational turns. All nodes whose importance score
falls below 0.10 are pruned, together with every edge incident to those nodes. Additional details are provided in Appendix~\ref{appendix:just}.

\section{Experiments}
\label{sec:experiments}
This section presents two experiments designed to evaluate (1) whether a knowledge graph retrieval
mechanism improves answer accuracy over a flat vector baseline, and (2) whether the proposed 
forgetting module can reduce graph storage without degrading retrieval quality. In all experiments,
the underlying language model is held constant and only the retrieval and memory mechanisms are varied.

\subsection{Dataset}

Both experiments use the LongMemEval benchmark \cite{wu2025longmemeval}, a dataset designed to
evaluate long-term conversational memory in dialogue systems. The benchmark consists of 500 questions, each paired with a multi-session conversation history, referred to as the \emph{haystack}, which contains the information required to answer the question. Haystack sessions span real historical dates covering approximately 33 months
(June 2021 -- February 2024), with each question drawing on between one and several sessions.

Questions are categorized into six types that probe distinct memory demands. \textit{Single-session (user)} 
questions ask about facts the user stated directly, such as personal attributes or past events. 
\textit{Single-session (assistant)} questions require recalling specific information the assistant
provided in a prior turn, such as a recommendation or a factual explanation. \textit{Single-session (preference)}
questions target implicit or explicit user preferences expressed in conversation. \textit{Knowledge-update}
questions test whether the system correctly tracks values that changed across sessions, favoring
the most recent statement over earlier ones. \textit{Multi-session} questions require aggregating
information spread across two or more separate sessions. \textit{Temporal-reasoning} questions demand
ordering events or computing time intervals from information embedded in the haystack.

\subsection{Experiment 1: Retrieval Quality and Knowledge Retention}
\label{sec:exp1}

The goal of Experiment 1 is to determine whether structuring conversational memory as a knowledge graph improves
retrieval quality over a flat vector baseline. For each of the 500 questions, a fresh knowledge graph is built from
that question's haystack sessions alone, then used to answer the question. The baseline stores the same haystack turns
as raw text chunks in a flat vector store and retrieves the top-5 most similar chunks at query time. Both systems use the
same language model for answer generation, and neither has access to information outside the question's own haystack.

Performance is measured using token-level precision and F1-score, and an LLM-as-judge correctness score
(binary, averaged across questions).

\begin{table}[h]
\caption{Experiment 1 results: Graph RAG vs.\ Baseline RAG on LongMemEval.
J = LLM-judge; Single = single-session.}
\centering
\setlength{\tabcolsep}{3pt}
\renewcommand{\arraystretch}{1.5}
\begin{tabular}{lcccccccc}
\hline
 & \multicolumn{4}{c}{\textbf{Graph RAG (Ours)}} & \multicolumn{4}{c}{\textbf{Baseline RAG}} \\
\cmidrule(lr){2-5} \cmidrule(lr){6-9}
\textbf{Type} & \textbf{n} & \textbf{P} & \textbf{F1} & \textbf{J} & \textbf{n} & \textbf{P} & \textbf{F1} & \textbf{J} \\
\hline
Single (user)       & 70  & 0.743 & 0.737 & 0.771 & 70  & 0.823 & 0.819 & 0.929 \\
Single (asst.)      & 56  & 0.680 & 0.575 & 0.607 & 56  & 0.916 & 0.774 & 0.911 \\
Single (pref.)      & 30  & \textbf{0.312} & 0.083 & 0.233 & 30  & 0.274 & 0.114 & 0.367 \\
Know.-update        & 78  & 0.507 & 0.456 & 0.513 & 78  & 0.553 & 0.511 & 0.590 \\
Multi-session       & 133 & \textbf{0.384} & 0.326 & 0.398 & 133 & 0.381 & 0.342 & 0.436 \\
Temporal           & 133 & \textbf{0.468} & 0.328 & \textbf{0.293} & 133 & 0.416 & 0.334 & 0.278 \\
\hline
\textbf{Overall}    & \textbf{500} & \textbf{0.505} & \textbf{0.417} & \textbf{0.454} & \textbf{500} & \textbf{0.532} & \textbf{0.468} & \textbf{0.536} \\
\hline
\end{tabular}
\vspace{3pt}

\label{tab:exp1}
\end{table}

\subsection{Experiment 2: Long-Term Memory Efficiency Under Forgetting}
The goal of Experiment 2 is to determine whether the forgetting module can compress the knowledge graph without
degrading its retrieval quality. A single persistent graph is built by ingesting the haystack sessions for all 500
questions in a sequence, simulating a long-running deployment in which a system accumulates memory across many independent
interactions. Two variants are compared: a \textit{no-forgetting} graph that retains every node, and a \textit{forgetting}
graph to which the forgetting module is applied once ingestion is complete. Both variants then answer all 500 benchmark
questions, and their retrieval quality and storage footprints are compared.

\begin{table}[h]
\caption{Experiment 2 results: storage and retrieval quality for the persistent graph 
with and without forgetting.}
\centering
\resizebox{0.75\linewidth}{!}{%
\begin{tabular}{lccccc}
\hline
\textbf{Variant} & \textbf{Nodes} & \textbf{Edges} & \textbf{Size} & \textbf{F1} & \textbf{Judge} \\
\hline
No-forgetting  & 27,021 & 46,538 & 440.6 MB & 0.292 & 0.300 \\
Forgetting     & 24,368 & 43,978 & 398.6 MB & 0.293 & 0.284 \\
\hline
\textbf{Change} & $-$9.8\% & $-$5.5\% & $-$9.5\% & $+$0.001 & $-$0.016 \\
\hline
\end{tabular}%
}
\vspace{3pt}
\label{tab:exp2}
\end{table}

\section{Discussion and Limitations}

The experiments reveal two complementary findings about graph-based long-term memory. First, representing conversational memory as a knowledge graph does not uniformly improve retrieval over a flat vector store. Second, once memory is represented as a persistent graph, selective forgetting can substantially reduce its size while largely preserving retrieval quality.

In Experiment~1, the baseline RAG system outperforms Graph RAG overall, achieving a token F1 of 0.468 compared with 0.417 and an LLM-judge accuracy of 0.536 compared with 0.454. However, performance varies across question types, indicating that the usefulness of graph structure depends on the type of information being retrieved.

Graph RAG achieves its only improvement on the LLM-judge metric for \textit{temporal-reasoning} questions (0.293 vs.\ 0.278). This task is naturally aligned with a graph representation: events can be represented as typed nodes with temporal attributes and connected to the entities that participate in them, allowing retrieval to preserve relational and temporal structure. In contrast, flat chunk retrieval does not explicitly represent event participants, ordering, or temporal relationships.

The largest performance deficit occurs for \textit{single-session (assistant)} questions (judge: 0.607 vs.\ 0.911). These questions often require recalling a specific recommendation or factual statement from a previous assistant response. The flat baseline can retrieve the original assistant turn verbatim, whereas graph extraction decomposes the turn into entities and relationships. In doing so, it may lose information about which item or statement was specifically emphasized. This highlights an important limitation of extraction-based memory representations: structured abstraction can improve relational organization while simultaneously discarding information required for precise or verbatim recall.

Graph RAG also underperforms on \textit{knowledge-update} questions (F1: 0.456 vs.\ 0.511). Inspection of failures indicates that the current conflict-resolution policy can retain an earlier attribute value instead of replacing it with a more recent value when no explicit confidence score is available. A last-write-wins policy for appropriate factual and numeric attributes could therefore improve performance on knowledge-update tasks. Smaller deficits on \textit{multi-session} and \textit{single-session (user)} questions appear to arise from related extraction and retrieval effects. Although graph structure can support cross-session entity linking, imperfect extraction and entity merging introduce retrieval noise, while concise facts that are directly preserved in raw text may be abstracted during graph construction.

Experiment~2 examines a different property of the memory system: whether accumulated graph memory can be reduced without substantially degrading retrieval. Applying the forgetting mechanism removes 2,653 nodes (9.8\%) and 2,560 edges (5.5\%), reducing the graph size from 440.6~MB to 398.6~MB, a 9.5\% reduction. Token-level F1 remains nearly unchanged, increasing from 0.292 to 0.293, while LLM-judge accuracy decreases from 0.300 to 0.284.

Absolute retrieval performance in Experiment~2 is lower than in Experiment~1 because all 500 haystacks are merged into a single persistent graph, introducing cross-conversation retrieval interference. The purpose of this experiment is therefore not to maximize retrieval accuracy, but to compare the same persistent-memory setting with and without forgetting.

The nodes removed by the forgetting mechanism fall below the importance threshold after the simulated conversation period and are characterized by low re-reference frequency, limited structural connectivity, and reduced recency. Their removal has little effect on token-level F1, suggesting that the importance function preferentially removes peripheral information that contributes relatively little to retrieval. The reduction in LLM-judge accuracy, however, indicates that some pruned information can still contribute to correct answers, highlighting a trade-off between memory efficiency and information retention.

Taken together, these results suggest that graph structure alone is not sufficient to improve long-term conversational memory. Its benefits are strongest when relationships and temporal structure are important, whereas flat text retrieval remains advantageous for precise or verbatim recall. At the same time, explicit retention mechanisms provide a practical way to control the growth of persistent memory. Effective long-term memory systems may therefore benefit from combining structured representations with stronger update policies, selective retention, and mechanisms that preserve access to information for which verbatim context remains important.

\section{Conclusion}
This study demonstrates that structuring conversational memory as a knowledge graph introduces both benefits and limitations. While relational representations can support reasoning over temporally and semantically connected information, they also incur information loss that negatively impacts tasks requiring precise or verbatim recall. These results indicate that improvements in memory systems cannot be achieved through representation alone. Instead, performance depends critically on how information is extracted, updated, and retained over time.

The proposed forgetting module contributes a retention mechanism whose cost
we can bound: pruning the low-importance tail of a 27{,}021-node store
removed 9.8\% of nodes and 9.5\% of bytes, and a paired bootstrap over 500
questions detects no significant change in any of the four metrics
(Table~\ref{tab:paired}).
 Overall, the findings suggest that effective long-term memory systems should combine structured representations with stronger update mechanisms and selective retention strategies, rather than relying on a single approach in isolation.

\section*{Author Contributions}

Sourena Khanzadeh conceived the research idea and formulated the initial
research direction. Theo Rusu was primarily responsible for the
implementation and experimental execution. Manar Alalfi supervised the
research and provided technical and academic guidance.

\bibliographystyle{plainnat}
\bibliography{references}

@article{zhang2024survey,
   title={A survey on the memory mechanism of large language model-based agents},
  author={Zhang, Zeyu and Dai, Quanyu and Bo, Xiaohe and Ma, Chen and Li, Rui and Chen, Xu and Zhu, Jieming and Dong, Zhenhua and Wen, Ji-Rong},
  journal={ACM Transactions on Information Systems},
  volume={43},
  number={6},
  pages={1--47},
  year={2025},
  publisher={ACM New York, NY}
}

@inproceedings{xu2022beyond,
   title={Beyond goldfish memory: Long-term open-domain conversation},
  author={Xu, Jing and Szlam, Arthur and Weston, Jason},
  booktitle={Proceedings of the 60th annual meeting of the association for computational linguistics (volume 1: long papers)},
  pages={5180--5197},
  year={2022}
}

@inproceedings{zhong2024memorybank,
title={Memorybank: Enhancing large language models with long-term memory},
  author={Zhong, Wanjun and Guo, Lianghong and Gao, Qiqi and Ye, He and Wang, Yanlin},
  booktitle={Proceedings of the AAAI conference on artificial intelligence},
  volume={38},
  number={17},
  pages={19724--19731},
  year={2024}
}

@inproceedings{park2023generative,
  title={Generative agents: Interactive simulacra of human behavior},
  author={Park, Joon Sung and O'Brien, Joseph and Cai, Carrie Jun and Morris, Meredith Ringel and Liang, Percy and Bernstein, Michael S},
  booktitle={Proceedings of the 36th annual acm symposium on user interface software and technology},
  pages={1--22},
  year={2023}
}

@inproceedings{lewis2020retrieval,
  title={Retrieval-augmented generation for knowledge-intensive nlp tasks},
  author={Lewis, Patrick and Perez, Ethan and Piktus, Aleksandra and Petroni, Fabio and Karpukhin, Vladimir and Goyal, Naman and K{\"u}ttler, Heinrich and Lewis, Mike and Yih, Wen-tau and Rockt{\"a}schel, Tim and others},
  journal={Advances in neural information processing systems},
  volume={33},
  pages={9459--9474},
  year={2020}
}

@inproceedings{lee2024readagent,
 title={A human-inspired reading agent with gist memory of very long contexts},
  author={Lee, Kuang-Huei and Chen, Xinyun and Furuta, Hiroki and Canny, John and Fischer, Ian},
  journal={arXiv preprint arXiv:2402.09727},
  year={2024}
}

@article{ji2022survey,
  title={A survey on knowledge graphs: Representation, acquisition, and applications},
  author={Ji, Shaoxiong and Pan, Shirui and Cambria, Erik and Marttinen, Pekka and Yu, Philip S},
  journal={IEEE transactions on neural networks and learning systems},
  volume={33},
  number={2},
  pages={494--514},
  year={2021},
  publisher={IEEE}
}

@article{wang2024forgetting,
  title={A comprehensive survey of forgetting in deep learning beyond continual learning},
  author={Wang, Zhenyi and Yang, Enneng and Shen, Li and Huang, Heng},
  journal={IEEE Transactions on Pattern Analysis and Machine Intelligence},
  volume={47},
  number={3},
  pages={1464--1483},
  year={2024},
  publisher={IEEE}
}

@inproceedings{maharana2024evaluating,
  title={Evaluating very long-term conversational memory of llm agents},
  author={Maharana, Adyasha and Lee, Dong-Ho and Tulyakov, Sergey and Bansal, Mohit and Barbieri, Francesco and Fang, Yuwei},
  booktitle={Proceedings of the 62nd Annual Meeting of the Association for Computational Linguistics (Volume 1: Long Papers)},
  pages={13851--13870},
  year={2024}
}

@inproceedings{wu2025longmemeval,
   title={Longmemeval: Benchmarking chat assistants on long-term interactive memory},
  author={Wu, Di and Wang, Hongwei and Yu, Wenhao and Zhang, Yuwei and Chang, Kai-Wei and Yu, Dong},
  journal={arXiv preprint arXiv:2410.10813},
  year={2024}
}

@inproceedings{sukhbaatar2015end,
  title={End-to-end memory networks},
  author={Sukhbaatar, Sainbayar and Weston, Jason and Fergus, Rob and others},
  journal={Advances in neural information processing systems},
  volume={28},
  year={2015}
}

@article{graves2014neural,
  title={Neural turing machines},
  author={Graves, Alex and Wayne, Greg and Danihelka, Ivo},
  journal={arXiv preprint arXiv:1410.5401},
  year={2014}
}

@book{ebbinghaus1885memory,
  title={A contribution to experimental psychology},
  author={Ebbinghaus, Hermann},
  journal={New York, NY: Teachers College, Columbia University},
  year={1913}
}

@article{mccloskey1989catastrophic,
  title={Catastrophic interference in connectionist networks: The sequential learning problem},
  author={McCloskey, Michael and Cohen, Neal J},
  booktitle={Psychology of learning and motivation},
  volume={24},
  pages={109--165},
  year={1989},
  publisher={Elsevier}
}

@article{kirkpatrick2017overcoming,
 title={Overcoming catastrophic forgetting in neural networks},
  author={Kirkpatrick, James and Pascanu, Razvan and Rabinowitz, Neil and Veness, Joel and Desjardins, Guillaume and Rusu, Andrei A and Milan, Kieran and Quan, John and Ramalho, Tiago and Grabska-Barwinska, Agnieszka and others},
  journal={Proceedings of the national academy of sciences},
  volume={114},
  number={13},
  pages={3521--3526},
  year={2017},
  publisher={National Academy of Sciences}
}

@inproceedings{bourtoule2021machine,
      title={Machine Unlearning}, 
      author={Lucas Bourtoule and Varun Chandrasekaran and Christopher A. Choquette-Choo and Hengrui Jia and Adelin Travers and Baiwu Zhang and David Lie and Nicolas Papernot},
      year={2020},
      eprint={1912.03817},
      archivePrefix={arXiv},
      primaryClass={cs.CR},
      url={https://arxiv.org/abs/1912.03817}, 
}

@inproceedings{karpukhin2020dense,
title={Dense passage retrieval for open-domain question answering},
  author={Karpukhin, Vladimir and Oguz, Barlas and Min, Sewon and Lewis, Patrick and Wu, Ledell and Edunov, Sergey and Chen, Danqi and Yih, Wen-tau},
  booktitle={Proceedings of the 2020 conference on empirical methods in natural language processing (EMNLP)},
  pages={6769--6781},
  year={2020}
}

@inproceedings{guu2020realm,
  title={Retrieval augmented language model pre-training},
  author={Guu, Kelvin and Lee, Kenton and Tung, Zora and Pasupat, Panupong and Chang, Mingwei},
  booktitle={International conference on machine learning},
  pages={3929--3938},
  year={2020},
  organization={PMLR}
}

@inproceedings{borgeaud2022improving,
  title={Improving language models by retrieving from trillions of tokens},
  author={Borgeaud, Sebastian and Mensch, Arthur and Hoffmann, Jordan and Cai, Trevor and Rutherford, Eliza and Millican, Katie and Van Den Driessche, George Bm and Lespiau, Jean-Baptiste and Damoc, Bogdan and Clark, Aidan and others},
  booktitle={International conference on machine learning},
  pages={2206--2240},
  year={2022},
  organization={PMLR}
}

@inproceedings{shuster2021retrieval,
title={Retrieval augmentation reduces hallucination in conversation},
  author={Shuster, Kurt and Poff, Spencer and Chen, Moya and Kiela, Douwe and Weston, Jason},
  booktitle={Findings of the Association for Computational Linguistics: EMNLP 2021},
  pages={3784--3803},
  year={2021}
}

@inproceedings{asai2024selfrag,
  title={Self-rag: Learning to retrieve, generate, and critique through self-reflection},
  author={Asai, Akari and Wu, Zeqiu and Wang, Yizhong and Sil, Avi and Hajishirzi, Hannaneh},
  booktitle={International conference on learning representations},
  volume={2024},
  pages={9112--9141},
  year={2024}
}

@article{luo2024chatqa,
 title={Chatqa: Surpassing gpt-4 on conversational qa and rag},
  author={Liu, Zihan and Ping, Wei and Roy, Rajarshi and Xu, Peng and Lee, Chankyu and Shoeybi, Mohammad and Catanzaro, Bryan},
  journal={Advances in Neural Information Processing Systems},
  volume={37},
  pages={15416--15459},
  year={2024}
}

@article{pan2024unifying,
   title={Unifying large language models and knowledge graphs: A roadmap},
  author={Pan, Shirui and Luo, Linhao and Wang, Yufei and Chen, Chen and Wang, Jiapu and Wu, Xindong},
  journal={IEEE Transactions on Knowledge and Data Engineering},
  volume={36},
  number={7},
  pages={3580--3599},
  year={2024},
  publisher={IEEE}
}

@inproceedings{baek2023knowledge,
   title={Knowledge-augmented language model prompting for zero-shot knowledge graph question answering},
  author={Baek, Jinheon and Aji, Alham Fikri and Saffari, Amir},
  booktitle={Proceedings of the 1st Workshop on Natural Language Reasoning and Structured Explanations (NLRSE)},
  pages={78--106},
  year={2023}
}

@inproceedings{sun2024thinkongraph,
 title={Think-on-graph: Deep and responsible reasoning of large language model on knowledge graph},
  author={Sun, Jiashuo and Xu, Chengjin and Tang, Lumingyuan and Wang, Saizhuo and Lin, Chen and Gong, Yeyun and Ni, Lionel and Shum, Heung-Yeung and Guo, Jian},
  booktitle={International Conference on Learning Representations},
  volume={2024},
  pages={3868--3898},
  year={2024}
}

@article{edge2024local,
 title={From local to global: A graph rag approach to query-focused summarization},
  author={Edge, Darren and Trinh, Ha and Cheng, Newman and Bradley, Joshua and Chao, Alex and Mody, Apurva and Truitt, Steven and Metropolitansky, Dasha and Ness, Robert Osazuwa and Larson, Jonathan},
  journal={arXiv preprint arXiv:2404.16130},
  year={2024}
}

@inproceedings{gutierrez2024hipporag,
title={Hipporag: Neurobiologically inspired long-term memory for large language models},
  author={Guti{\'e}rrez, Bernal J and Shu, Yiheng and Gu, Yu and Yasunaga, Michihiro and Su, Yu},
  journal={Advances in neural information processing systems},
  volume={37},
  pages={59532--59569},
  year={2024}
}

@inproceedings{packer2024memgpt,
  title={Memgpt: Towards llms as operating systems},
  author={Packer, Charles and Wooders, Sarah and Lin, Kevin and Fang, Vivian and Patil, Shishir G and Stoica, Ion and Gonzalez, Joseph E},
  journal={arXiv preprint arXiv:2310.08560},
  year={2023}
}

@article{liu2023think,
  title={Think-in-memory: Recalling and post-thinking enable llms with long-term memory},
  author={Liu, Lei and Yang, Xiaoyan and Shen, Yue and Hu, Binbin and Zhang, Zhiqiang and Gu, Jinjie and Zhang, Guannan},
  journal={arXiv preprint arXiv:2311.08719},
  year={2023}
}

@article{wang2024selfcontrolled,
   title={Scm: Enhancing large language model with self-controlled memory framework},
  author={Wang, Bing and Liang, Xinnian and Yang, Jian and Huang, Hui and Wu, Zhenhe and Wu, ShuangZhi and Ma, Zejun and Li, Zhoujun},
  booktitle={International Conference on Database Systems for Advanced Applications},
  pages={188--203},
  year={2025},
  organization={Springer}
}

@article{modarressi2024memllm,
  title={Memllm: Finetuning llms to use an explicit read-write memory},
  author={Modarressi, Ali and K{\"o}ksal, Abdullatif and Imani, Ayyoob and Fayyaz, Mohsen and Sch{\"u}tze, Hinrich},
  journal={arXiv preprint arXiv:2404.11672},
  year={2024}
}

@article{gao2023retrieval,
  title={Retrieval-augmented generation for large language models: A survey},
  author={Gao, Yunfan and Xiong, Yun and Gao, Xinyu and Jia, Kangxiang and Pan, Jinliu and Bi, Yuxi and Dai, Yi and Sun, Jiawei and Wang, Meng and Wang, Haofen},
  journal={arXiv preprint arXiv:2312.10997},
  year={2023}
}

@article{chhikara2025mem0,
   title={Mem0: Building production-ready ai agents with scalable long-term memory},
  author={Chhikara, Prateek and Khant, Dev and Aryan, Saket and Singh, Taranjeet and Yadav, Deshraj},
  journal={arXiv preprint arXiv:2504.19413},
  year={2025}
}

@article{peng2023knowledge,
      title={Knowledge Graphs: Opportunities and Challenges}, 
      author={Ciyuan Peng and Feng Xia and Mehdi Naseriparsa and Francesco Osborne},
      year={2023},
      eprint={2303.13948},
      archivePrefix={arXiv},
      primaryClass={cs.AI},
      url={https://arxiv.org/abs/2303.13948}, 
}

@article{wei2026fademem,
  title={Fademem: Biologically-inspired forgetting for efficient agent memory},
  author={Wei, Lei and Dong, Xu and Peng, Xiao and Xie, Niantao and Wang, Bin},
  booktitle={ICASSP 2026-2026 IEEE International Conference on Acoustics, Speech and Signal Processing (ICASSP)},
  pages={4011--4015},
  year={2026},
  organization={IEEE}
}

\appendix

\section{Appendix}

\subsection{Note to Reviewers on Experimental Scope and AI Use}

\paragraph{Budget Constraints.}
This work was carried out under a fixed compute and API budget on the
consumer-grade workstation described above. We state the resulting scope
limits explicitly so that our claims are read at the right granularity.
Every configuration we evaluate requires re-ingesting the haystack sessions
for all 500 LongMemEval questions, which costs one extraction call per
conversational turn, plus one generation and one judge call per question.
A single evaluated configuration is therefore expensive, and the budget
admitted a small number of complete runs rather than a sweep. We chose to
spend it on four full runs (Experiment 1 treatment and control, Experiment 2
treatment and control) at $n=500$ with temperature $=0$, and to report
paired bootstrap intervals over those runs, rather than on a larger number
of partially evaluated configurations.

Accordingly, our findings should be read as characterising this
extraction-based graph memory pipeline at this model scale, not
graph-structured memory in general. Given additional budget, our order of
priority would be: a matched-compression control that prunes the same
fraction of nodes at random, to isolate the contribution of the importance
function; a second benchmark; and a stronger extraction model.

\paragraph{Use of Generative AI.}
We distinguish two uses of language models in this work. First, as components of the method itself: GPT-4o-mini performs knowledge-graph extraction and answer generation, and serves as the LLM judge, as described in Sections 3.3 and 4.1. Second, as authoring tools. In the latter role, we used gpt5.6 sol to draft prose in the appendix and to assist with literature
search, and Opus 4.8 to assist with implementing the experimental pipeline.
The main technical content, including the experimental design, the analyses, and the interpretation of results, is the authors' own; model assistance on those sections was limited to grammar and formatting. All AI-assisted output was reviewed by the authors, all cited references were checked against their
original sources, and the authors take full responsibility for the content of the paper.

\subsection{Justification for parameter values chosen for the experiments and methodology.}
\label{appendix:just}
Table~\ref{tab:params} lists every free parameter of the system together with its
setting and the basis on which it was chosen. We distinguish three cases:
values selected empirically on a small probe set (E), values fixed \emph{a
priori} from a budget or cost constraint (B), and values fixed by convention or
by symmetry with the baseline (C). We did not perform a full sweep over the
retention parameters; the consequences of this are discussed at the end of this
section.

\begin{table}[h]
\centering
\small
\caption{Parameters, settings, and basis for selection. E = empirical probe,
B = budget or cost constraint, C = convention or symmetry with the baseline.}
\label{tab:params}
\begin{tabular}{llc}
\toprule
Parameter & Value & Basis \\
\midrule
De-duplication cosine threshold $\tau_{\text{dedup}}$ & 0.92 & E \\
Retrieval cosine floor $\tau_{\text{ret}}$ & 0.75 & E \\
Retrieval roots (top-$k$) & 5 & C \\
Subgraph expansion depth & 2 hops & B \\
Subgraph node cap & 15 & B \\
Forgetting interval & 400 turns & B \\
Pruning threshold $s_{\min}$ & 0.10 & B \\
Recency half-life & 90 days & C \\
Turn-decay half-life & 1{,}000 turns & C \\
Scoring weights $(w_r, w_f, w_c, w_t)$ & $(0.35, 0.25, 0.20, 0.20)$ & C \\
\bottomrule
\end{tabular}
\end{table}

\textbf{De-duplication threshold.} Selected by the probe procedure described in
Appendix~\ref{appendix:algorithms}. The value is deliberately conservative
because the two error modes are not symmetric: a false merge collapses two
distinct entities irreversibly and corrupts every edge incident to them, whereas
a missed merge only leaves redundant nodes that later de-duplication passes or
the retrieval stage can still surface. We therefore accepted a higher
false-negative rate in exchange for a low false-merge rate.

\textbf{Retrieval cosine floor.} Descriptor embeddings produced by
\texttt{nomic-embed-text} are anisotropic, so cosine similarity between
unrelated short descriptors does not concentrate near zero. A floor of 0.75
admits paraphrases and partial mentions of the same entity while excluding
nodes that are merely topically adjacent. This parameter is not load-bearing:
because candidates are subsequently ranked and truncated to the top 5, the
floor only affects queries for which fewer than five nodes clear it, in which
case the system correctly retrieves a smaller context rather than padding it
with unrelated nodes.

\textbf{Number of retrieval roots.} Set to 5 to match the number of chunks
retrieved by the flat vector baseline (Section~\ref{sec:exp1}), so that the two
systems are compared at an equal candidate-generation budget and any difference
in performance is attributable to the representation rather than to the number
of retrieval hits.

\textbf{Expansion depth and node cap.} One hop from a seed node returns only its
immediate neighbours, which for most seeds is the set of attributes attached to
a single entity and therefore adds little beyond the seed itself. Three or more
hops expand super-linearly in a merged graph, and inspection showed that nodes
at that distance are typically related to the seed through a hub entity rather
than through any relation relevant to the query. Two hops is thus the smallest
depth that supports the relational and multi-session cases the graph
representation is intended to serve. 

\textbf{Forgetting interval and pruning threshold.} Scoring is $O(N + M)$ in the
size of the graph, so invoking it rarely amortizes its cost across many
conversational turns; 400 turns is also long enough for the recency and
turn-decay terms to separate nodes that are genuinely dormant from nodes that
happen not to have been accessed recently. The threshold of 0.10 was set to be
conservative, targeting only the tail of the score distribution rather than a
fixed compression ratio. This choice determines the operating point reported in
Experiment~2 and is the reason the reported reduction is approximately 10\%
rather than a larger figure.

\textbf{Half-lives and scoring weights.} The 90-day recency half-life is set
relative to the temporal span of the LongMemEval haystacks, which cover roughly
33 months of real timestamps; a substantially shorter half-life would saturate
the recency term for nearly all nodes, and a substantially longer one would
flatten it. The 1{,}000-turn decay half-life plays the same role with respect to
ingestion order. The weights were fixed \emph{a priori} and were not tuned: they
encode a prior, drawn from the memory-stream and forgetting-curve literature
\citep{park2023generative, zhong2024memorybank}, that access-driven signals
(recency, frequency) should dominate structural and age-based ones.

\textbf{Limitation.} The retention parameters $(w_r, w_f, w_c, w_t)$,
$s_{\min}$, and the forgetting interval were not swept, so Experiment~2
characterises a single point on the compression--quality trade-off rather than
the curve. We also do not isolate the contribution of the individual scoring
components; establishing that the four-term importance function outperforms a
simpler retention rule at matched compression is left to future work.

\paragraph{Compute Environment.}
All experiments were conducted on a local workstation equipped with an
AMD Radeon RX 7800 XT GPU, 16\,GB of system RAM, and an Intel Core
i5-9400F CPU. This configuration was used for local execution of the
memory pipeline, graph operations, embedding-related workloads and
we utilized OpenAI model, mainly (gpt4o-mini) for API calls.

\subsection{Role-Aware Extraction Prompt}
\label{app:prompt-role-aware}

This role-aware variant explicitly distinguishes user and assistant
turns and specifies which assistant-provided facts should be retained.

\begin{promptbox}{Role-Aware Extraction Prompt}
Extract a knowledge graph from a conversation turn.
Return strict JSON only -- no markdown, no preamble.

Each turn is prefixed with its speaker role:
  [Role: user]      -- the human speaking directly
  [Role: assistant] -- the AI assistant responding

ROLE HANDLING:
...
\end{promptbox}

\subsection{Sample Knowledge-Graph Extraction Prompt}
\label{app:sample-prompt}

The following example illustrates the structure of the extraction prompts
used throughout our experiments.

\begin{promptbox}{Sample Knowledge-Graph Extraction Prompt}
You are a knowledge graph extraction engine. Given a single
conversational message, extract all relevant nodes, edges, and
attributes and return the result as strict JSON.

NODES:
Represent discrete entities mentioned or implied in the message,
such as people, organizations, locations, skills, goals, events,
preferences, and artifacts.

EDGES:
Represent directed relationships between nodes. Use precise,
domain-relevant relationship labels whenever possible.

ATTRIBUTES:
Represent additional properties associated with a node or edge,
including quantities, dates, durations, frequencies, or other
qualifying information.

REQUIREMENTS:
1. Extract every explicitly stated fact.
2. Do not introduce information that is not supported by the input.
3. Use only the predefined node labels.
4. Return valid JSON only.
5. Do not include explanations, markdown, or additional commentary.

OUTPUT FORMAT:
{
  "nodes": {
    "<node_id>": {
      "label": "<node_label>",
      "title": "<short_title>",
      "content": "<description>",
      "attributes": {}
    }
  },
  "edges": {
    "<edge_id>": {
      "source": "<source_node>",
      "target": "<target_node>",
      "relationship": "<relationship>",
      "attributes": {}
    }
  }
}

USER MESSAGE:
<conversation turn>
\end{promptbox}

\subsection{Algorithms}
\label{appendix:algorithms}

Algorithm~\ref{alg:ingestion} summarizes the memory ingestion pipeline. Given a conversational turn, the system first extracts a structured set of nodes and edges and embeds each extracted node. Candidate nodes are matched against existing memory using exact title matching followed by semantic similarity. Matches above the deduplication threshold $\tau_{\mathrm{dedup}}$ are mapped to existing nodes, while unmatched entities are assigned new identifiers. For standard operation, the same message is additionally converted into retrieval entities, which are used to identify and expand a relevant graph subgraph that is serialized as contextual memory. The extracted nodes and edges are then written to persistent memory, and an optional forgetting pass is triggered according to the configured memory-maintenance policy.

\begin{algorithm}[t]
\caption{Knowledge-Graph Ingestion Pipeline}
\label{alg:ingestion}
\begin{algorithmic}[1]
\Require User message $m$, memory graph $G$
\Ensure Updated graph $G$ and retrieved context $\mathcal{C}$

\State $E \gets \textsc{Extract}(m)$
\Comment{Extract nodes and relations}

\State $Z \gets \textsc{Embed}(E.\text{nodes})$
\Comment{Embed extracted entities}

\ForAll{$n \in E.\text{nodes}$}
    \State $v \gets \textsc{Match}(n, Z_n, G)$
    \If{$v$ is sufficiently similar to $n$}
        \State $\textsc{Merge}(n,v,G)$
    \Else
        \State $\textsc{AddNode}(n,G)$
    \EndIf
\EndFor

\State $\textsc{AddRelations}(E.\text{edges},G)$

\State $Q \gets \textsc{ExtractQueryEntities}(m)$
\State $R \gets \textsc{RetrieveRelevantNodes}(Q,G)$
\State $S \gets \textsc{ExpandSubgraph}(R,G)$
\State $\mathcal{C} \gets \textsc{Serialize}(S)$

\If{\textsc{ForgettingTriggered}$(G)$}
    \State $G \gets \textsc{Forget}(G)$
\EndIf

\State \Return $(\mathcal{C},G)$
\end{algorithmic}
\end{algorithm}

\paragraph{Deduplication Threshold Selection.}
To identify an appropriate deduplication threshold, we constructed a sequence of prompts containing repeated references to the same underlying entities while varying the wording and contextual phrasing across prompts. This allowed us to evaluate how consistently semantically equivalent entities were merged as the similarity threshold changed. We then selected the threshold that provided the best trade-off between correctly merging duplicate entities and avoiding incorrect merges between distinct entities.

\subsection{Computational Complexity}
\label{app:complexity}

Let $N$ and $M$ be the number of nodes and edges in the memory graph, $d$ the
embedding dimension, $k$ the number of entities extracted from an incoming
message, $\ell$ the number of relations extracted with them, and $q$ the number
of entities extracted from a query. Table~\ref{tab:complexity} summarises the
cost of each stage; we exclude the internal cost of the LLM and embedding
calls, which depends on token lengths rather than on graph size.

\begin{table}[h]
\centering
\small
\caption{Per-stage complexity under exhaustive vector search. $T$ is the
forgetting interval (400 turns), and $V_S, E_S$ are the nodes and edges of the
retrieved subgraph.}
\label{tab:complexity}
\begin{tabular}{lll}
\toprule
Stage & Cost & Note \\
\midrule
Embed extracted entities & $O(kd)$ & \\
Title de-duplication & $O(1)$ per entity & hash index \\
Embedding de-duplication & $O(kNd)$ & linear scan \\
Graph write & $O(k+\ell)$ & indexed updates \\
Candidate retrieval & $O(qNd)$ & linear scan \\
Subgraph traversal & $O(N + M + |V_S| + |E_S|)$ & adjacency build \\
Forgetting & $O(N+M)$ every $T$ turns & amortized $O\!\left(\tfrac{N+M}{T}\right)$ \\
\bottomrule
\end{tabular}
\end{table}

Two exhaustive vector scans dominate, one at de-duplication and one at
retrieval, giving a worst-case per-turn cost of
\[
O\big((k+q)Nd + N + M\big) \;=\; O(Nd + M),
\]
since $k$, $q$, and the subgraph size are bounded by construction (the traversal
is capped at 15 nodes). The $N+M$ term arises only because the adjacency
representation is rebuilt at query time and would vanish if it were persisted
alongside the graph. Space complexity is $O(Nd + N + M)$, where $Nd$ accounts
for stored embeddings and $N+M$ for graph structure and metadata.

The linear-scan terms are therefore the only components that grow with memory
size, and both are incidental to the design: replacing the exhaustive search
with an approximate nearest-neighbour index would reduce the $O(Nd)$ factor
substantially, leaving the forgetting module as the mechanism that bounds $N$
itself.

\subsection{Statistical Significance of Experimental Results}
\label{app:statistical-significance}

We report the statistical significance of the results underlying our main claim, namely that the forgetting module handles long-term memory growth efficiently: it substantially reduces graph storage (Table~\ref{tab:exp2}) without a statistically significant loss in answer quality. All intervals below are computed post-hoc over the per-question results produced by our benchmark runs (Section~\ref{sec:experiments}); no additional model calls were made to compute them.

\paragraph{Setup.}
The factor of variability captured by every interval below is \emph{which questions were sampled from the fixed LongMemEval evaluation set}, i.e.\ we resample over questions; answer generation used temperature $=0$, so there is no additional decoding-stochasticity component to capture. Rows corresponding to a crashed or errored run were dropped before aggregation. For each condition and metric we report the mean together with the standard error of the mean (SEM, a 1-$\sigma$ interval, stated explicitly as such) and a 95\% confidence interval obtained from a nonparametric bootstrap (10,000 resamples with replacement over questions, percentile method), which makes no Normality assumption on the underlying metric distribution. For the binary judge-correctness metric, whose sampling distribution is a proportion bounded in $[0,1]$, we additionally report the 95\% Wilson score interval, which by construction cannot extend outside $[0,1]$; we prefer it over a symmetric interval for this metric to avoid the risk of implying out-of-range values.

\paragraph{Per-Condition Results.}
Table~\ref{tab:stat-significance-conditions} reports the mean $\pm$ SEM and the 95\% bootstrap confidence interval for each of the four experimental conditions.

\begin{table}[h]
\centering
\caption{Mean $\pm$ SEM (1$\sigma$) and 95\% bootstrap CI for each condition, $n=500$ questions per condition.}
\label{tab:stat-significance-conditions}
\resizebox{\linewidth}{!}{

\begin{tabular}{lcccc}
\toprule
Condition & F1 & Precision & Recall & Judge acc. \\
\midrule
Baseline RAG (per-question, Experiment 1 control) & $0.468 \pm 0.019$ & $0.532 \pm 0.019$ & $0.486 \pm 0.020$ & $0.536 \pm 0.022$ \\
Graph RAG (per-question, Experiment 1 treatment) & $0.417 \pm 0.019$ & $0.505 \pm 0.020$ & $0.412 \pm 0.019$ & $0.454 \pm 0.022$ \\
Persistent graph, no forgetting (Experiment 2 control) & $0.292 \pm 0.017$ & $0.363 \pm 0.018$ & $0.295 \pm 0.018$ & $0.300 \pm 0.021$ \\
Persistent graph, with forgetting (Experiment 2 treatment) & $0.293 \pm 0.017$ & $0.359 \pm 0.018$ & $0.294 \pm 0.018$ & $0.284 \pm 0.020$ \\
\bottomrule
\end{tabular}
}
\end{table}

\begin{table}[h]
\centering
\caption{95\% bootstrap confidence intervals corresponding to Table~\ref{tab:stat-significance-conditions}. For judge accuracy, the 95\% Wilson score interval is shown alongside the bootstrap CI.}
\label{tab:stat-significance-conditions-ci}
\resizebox{\linewidth}{!}{
\begin{tabular}{lcccc}
\toprule
Condition & F1 & Precision & Recall & Judge acc. \\
\midrule
Baseline RAG (per-question, Experiment 1 control) & $[0.432, 0.505]$ & $[0.496, 0.571]$ & $[0.447, 0.525]$ & $[0.492, 0.580]$ (Wilson: $[0.492, 0.579]$) \\
Graph RAG (per-question, Experiment 1 treatment) & $[0.381, 0.454]$ & $[0.467, 0.544]$ & $[0.375, 0.451]$ & $[0.410, 0.496]$ (Wilson: $[0.411, 0.498]$) \\
Persistent graph, no forgetting (Experiment 2 control) & $[0.260, 0.325]$ & $[0.328, 0.399]$ & $[0.261, 0.330]$ & $[0.260, 0.342]$ (Wilson: $[0.262, 0.342]$) \\
Persistent graph, with forgetting (Experiment 2 treatment) & $[0.260, 0.326]$ & $[0.323, 0.395]$ & $[0.259, 0.329]$ & $[0.244, 0.324]$ (Wilson: $[0.246, 0.325]$) \\
\bottomrule
\end{tabular}
}
\end{table}

\paragraph{Paired Comparisons.}
To assess whether the differences between paired conditions are statistically significant, we compute a paired nonparametric bootstrap over the per-question difference in each metric, matched by question ID between the two conditions being compared (10,000 resamples). We report the mean difference, its 95\% bootstrap CI, and a two-sided bootstrap $p$-value (twice the smaller tail fraction of resampled differences crossing zero, capped at 1). A comparison is marked significant at the $\alpha=0.05$ level when the 95\% CI on the difference excludes zero. Results are shown in Table~\ref{tab:stat-significance-contrasts}.

\begin{table}[h]
\centering
\caption{Paired bootstrap comparisons between conditions. A positive mean difference favors the first-named condition; $^{*}$ denotes significance at $\alpha=0.05$.}
\label{tab:stat-significance-contrasts}
\resizebox{\linewidth}{!}{

\begin{tabular}{lcccc}
\toprule
Comparison & F1 & Precision & Recall & Judge acc. \\
\midrule
Experiment 1: Graph RAG vs Baseline RAG & $-0.050$$^{*}$ $[-0.085, -0.016]$ & $-0.028$ $[-0.067, +0.011]$ & $-0.073$$^{*}$ $[-0.107, -0.039]$ & $-0.082$$^{*}$ $[-0.128, -0.034]$ \\
Experiment 2: Forgetting vs No forgetting & $+0.001$ $[-0.015, +0.016]$ & $-0.004$ $[-0.023, +0.014]$ & $-0.001$ $[-0.017, +0.016]$ & $-0.016$ $[-0.038, +0.006]$ \\
\bottomrule
\end{tabular}
}
\label{tab:paired}
\end{table}

\paragraph{Interpretation.}
None of the four metrics show a significant difference between the forgetting and no-forgetting conditions (Experiment 2), which is the key evidence for the no-quality-loss half of our main claim. The Graph~RAG vs.\ Baseline~RAG comparison (Experiment 1) is included for completeness but is not load-bearing for our main claim.

\end{document}